\documentclass[11pt]{article}
\usepackage[letterpaper,margin=1in]{geometry}
\usepackage[T1]{fontenc}
\usepackage[utf8]{inputenc}
\usepackage{lmodern}
\usepackage{graphicx}
\usepackage{amsmath}
\usepackage{booktabs}
\usepackage{microtype}
\usepackage{hyperref}
\usepackage{caption}
\hypersetup{hidelinks}
\newcommand{\XCOtwo}{XCO$_2$}
\newcommand{\COtwo}{CO$_2$}

\title{\textbf{GeoPrior-Mamba: Structured Process Priors with Mamba for Fine-Resolution \XCOtwo{} Reconstruction}}
\author{Zhao Meng$^{1}$, Yinan Cai$^{2}$, Siru Zhong$^{1}$, Juepeng Zheng$^{3*}$, Haohuan Fu$^{1*}$\\[0.4em]
\small $^{1}$Tsinghua Shenzhen International Graduate School, Tsinghua University, Shenzhen, China\\
\small $^{2}$National Supercomputing Center in Shenzhen, Shenzhen, China\\
\small $^{3}$School of Artificial Intelligence, Sun Yat-Sen University, Zhuhai, China\\
\small $^{*}$Corresponding authors: Haohuan Fu, Juepeng Zheng\\
\small haohuan@tsinghua.edu.cn; zhengjp8@mail.sysu.edu.cn; mzhaowl77@mail.tsinghua.edu.cn}
\date{}

\begin{document}
\maketitle

\begin{abstract}
Reconstructing fine-resolution column-averaged dry-air \COtwo{} (\XCOtwo{}) fields from sparse satellite observations requires models to infer spatial structure that is only weakly constrained by direct measurements. Existing learning-based methods typically treat environmental covariates as ordinary numerical inputs and must therefore learn heterogeneous source--sink relationships largely from sparse supervision. We introduce GeoPrior-Mamba, a multi-directional Mamba framework augmented with offline language-model-induced structured process priors. Rather than using a language model to predict \XCOtwo{}, we use it before training to organize relative process knowledge for biospheric uptake, ecosystem respiration, and anthropogenic emissions into deterministic prior tables. These priors are spatially instantiated using geographic, ecological, emission-related, and seasonal information and are adaptively injected into the reconstruction backbone through a lightweight knowledge adapter. Using OCO-2 observations from 2018--2020, GeoPrior-Mamba achieves an RMSE of 0.81 ppm and an R$^2$ of 0.93 on held-out observations, reducing RMSE by 48.2\% relative to CAMS background interpolation and by 3.1\% relative to Trans-XCO2 under the same evaluation protocol. Ablation experiments show a measurable contribution from the knowledge-prior branch and substantially faster convergence than the knowledge-free Mamba backbone. Independent TCCON evaluation further supports the consistency of the reconstructed fields with ground-based column \COtwo{} measurements. These results suggest that language models can provide a practical mechanism for constructing structured process priors when globally consistent process-response representations are difficult to obtain directly, while remaining outside the numerical prediction loop.
\end{abstract}

\section{Introduction}
Satellite measurements have greatly expanded the spatial coverage of atmospheric carbon dioxide observations, but they remain sparse and irregular in space and time. For column-averaged dry-air \COtwo{} (\XCOtwo{}), valid retrievals from missions such as OCO-2 are limited by cloud and aerosol contamination, surface reflectance, illumination geometry, orbital sampling, and retrieval quality control. Consequently, satellite observations alone do not provide spatially continuous fine-resolution \XCOtwo{} fields. (Eldering et al., 2017; Pandey et al., 2024; Yang et al., 2025)

Reconstructing \XCOtwo{} from incomplete observations is challenging because large parts of the target field are only weakly constrained by direct measurements. Existing statistical, assimilation-based, and learning-based methods address this problem in different ways. Data-driven methods are especially flexible because they can combine atmospheric background fields, meteorological variables, vegetation indices, emissions, geographic information, and temporal signals. However, these predictors are usually treated as ordinary numerical channels. The model must therefore infer relationships associated with biospheric uptake, ecosystem respiration, anthropogenic emissions, and their geographic and seasonal dependence largely from sparse satellite supervision. Representative reconstruction studies are discussed in Section 2.1 (Gao et al., 2023; He et al., 2022; Li et al., 2023; Wang et al., 2023; Zhang et al., 2023b).

Many of these relationships are already partly understood at the process level, but globally consistent and spatially complete representations of such process-response knowledge are not readily available from existing observational or model products. The challenge is therefore to organize dispersed scientific knowledge into a deterministic, computable form that is spatially aligned with gridded data and compatible with end-to-end learning. Large language models provide one possible mechanism for this organization. We therefore ask a different question: can language-model knowledge be converted offline into structured numerical priors that guide a geoscientific reconstruction model? (Zhang et al., 2024; Zscheischler et al., 2017)

We propose GeoPrior-Mamba, which separates knowledge organization from numerical prediction. An offline language model is used to construct fixed relative-activity priors for major \XCOtwo{} source--sink processes under different geographic and ecological conditions. These priors are then instantiated as pixel-wise fields and adaptively injected into a multi-directional Mamba reconstruction backbone. The language model is never used to predict \XCOtwo{} directly and is not called during model training or inference.

Our contributions are threefold:
\begin{enumerate}
\item We introduce a framework for converting offline language-model knowledge into deterministic, spatially instantiated process priors for sparse geoscientific reconstruction.
\item We develop GeoPrior-Mamba, which adaptively integrates these priors with a shared multi-directional Mamba backbone for daily global \XCOtwo{} reconstruction.
\item We evaluate the framework using OCO-2 observations from 2018--2020, showing improved held-out reconstruction relative to CAMS background interpolation and Trans-XCO2, measurable gains over a knowledge-free Mamba backbone, faster convergence, and broad consistency with independent TCCON observations.
\end{enumerate}

\section{Related Work}
\subsection{XCO2 Reconstruction from Sparse Observations}
Existing \XCOtwo{} reconstruction methods include statistical interpolation, data-assimilation products, and learning-based approaches that combine satellite retrievals with auxiliary environmental predictors. Recent learning-based methods improve spatial completeness and resolution, but they generally represent meteorological, vegetation, emission, and geographic information as ordinary numerical inputs. GeoPrior-Mamba is complementary to this line of work: its main distinction is not a new source of supervision, but an explicit mechanism for injecting structured source--sink process context into a sparse reconstruction model. Representative examples include statistical and ensemble reconstruction, self-supervised fusion, high-resolution deep learning, and recent global daily products (Gao et al., 2023; Guan et al., 2024; He et al., 2022; Li et al., 2022; Wang et al., 2023; 2025; Zhang et al., 2023b).

\subsection{Knowledge-Guided Scientific Machine Learning}
Knowledge-guided and physics-informed learning incorporate scientific structure through governing equations, simulation outputs, expert rules, constraints, or structured priors. Our setting differs from approaches that directly enforce physical equations because fine-resolution \XCOtwo{} reconstruction does not provide a complete closed-form process model at the target scale. Instead, we use structured relative-activity priors as soft contextual guidance and allow the neural network to learn when and where they are useful. (Raissi et al., 2019; Zscheischler et al., 2017)

\subsection{Language Models as Scientific Knowledge Sources}
Language models are increasingly used to organize scientific information, generate hypotheses, retrieve domain knowledge, or support scientific workflows. GeoPrior-Mamba uses a more constrained role: the language model is queried only offline to construct fixed numerical priors, and no language-model component is present during training or inference. This design reduces computational overhead and avoids making numerical predictions depend directly on stochastic language-model outputs. (Zhang et al., 2024)

\begin{figure}[t]
\centering
\includegraphics[width=\textwidth]{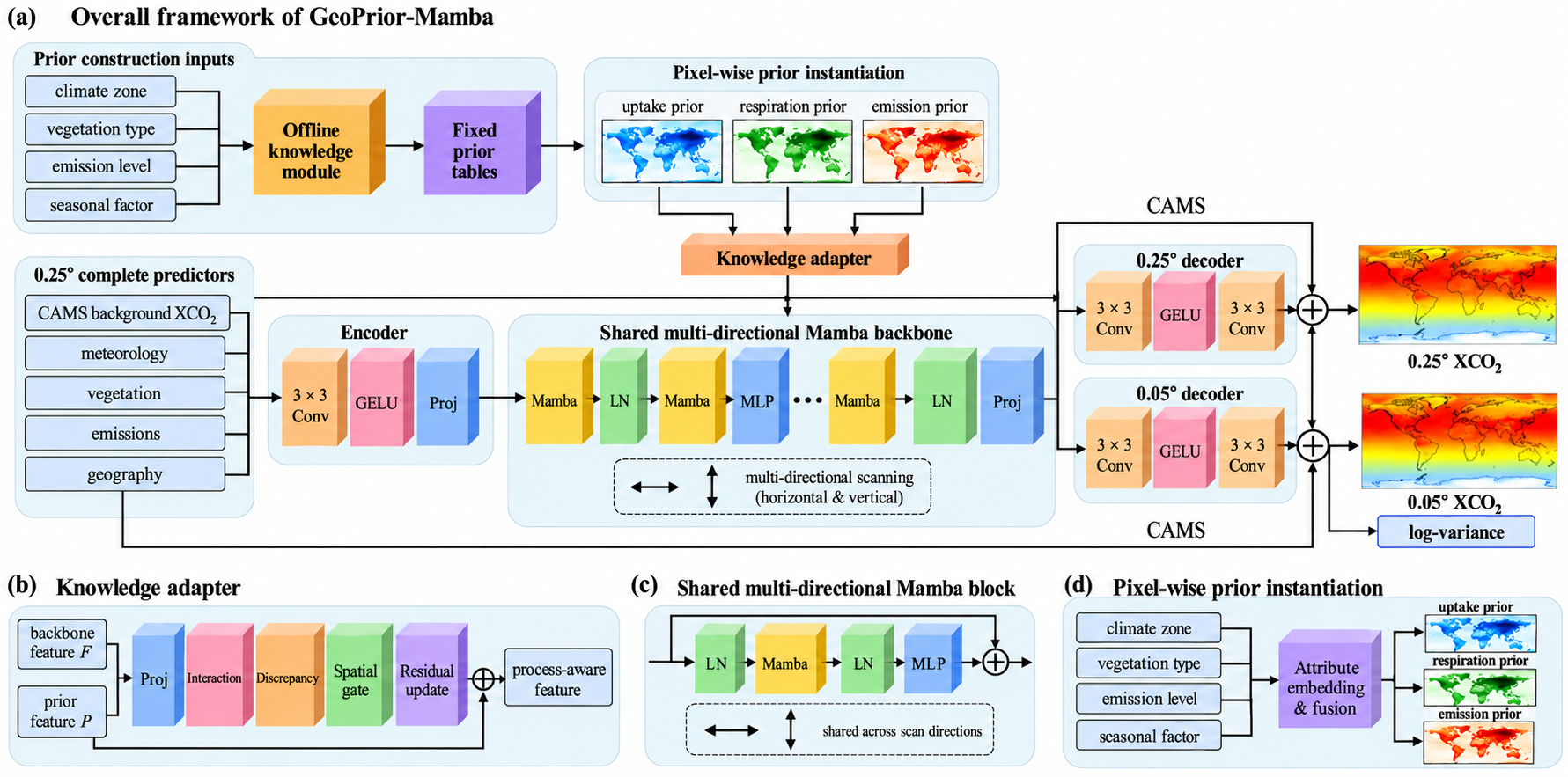}
\caption{Architecture of the proposed GeoPrior-Mamba reconstruction model.}
\label{fig:arch}
\end{figure}

\section{Problem Setup and Data}
We formulate \XCOtwo{} reconstruction as supervised learning with sparse observational constraints. The model receives complete gridded predictors on a daily 0.25$^\circ$ input grid and produces a spatially continuous \XCOtwo{} field on a 0.05$^\circ$ target grid. OCO-2 observations mapped to the 0.05$^\circ$ grid provide supervision only where valid retrievals are available; unobserved grid cells do not contribute to the supervised reconstruction loss.

The predictors include a CAMS \XCOtwo{} background field, ERA5 meteorological variables, MODIS vegetation indices, ODIAC fossil-fuel emissions, and static geographic attributes. CAMS provides the large-scale atmospheric background, while meteorological, vegetation, emission, and geographic variables provide information related to transport, surface conditions, biospheric activity, anthropogenic forcing, and spatial heterogeneity. Climate zone, vegetation type, and emission level are additionally used to instantiate the structured process priors described in Section 4. Independent TCCON observations are used only for external evaluation and are never used for model training.

Table S1 summarizes the datasets and their roles in the reconstruction framework. Data sources, preprocessing, and native resolutions are summarized in Section 3.

\section{Method}
\subsection{Framework Overview}
As shown in Figure~\ref{fig:arch}, GeoPrior-Mamba contains three components. First, an offline language-model-based knowledge module organizes process knowledge for biospheric uptake, ecosystem respiration, and anthropogenic emissions into fixed relative-activity priors. Second, these priors are mapped to each grid cell and date using geographic, ecological, emission-related, and seasonal attributes. Third, a knowledge adapter injects the resulting process fields into a multi-directional Mamba backbone that reconstructs spatially continuous \XCOtwo{}.

This design explicitly separates knowledge elicitation, prior instantiation, and numerical reconstruction. The language model is used only before training, while the reconstruction network remains fully deterministic during training and inference.

\subsection{Offline Construction of Structured Process Priors}
For biospheric processes, the language model is queried for combinations of climate zone and vegetation type to obtain candidate relative-activity scores for vegetation uptake and ecosystem respiration. These values are normalized process indicators rather than estimates of physical carbon flux. Anthropogenic forcing is represented by a separate prior indexed by emission level, while vegetation-dependent seasonal factors modulate the biospheric priors across reconstruction dates.

To reduce sensitivity to individual language-model outputs, multiple independent responses are generated for each attribute combination. Candidate values are screened using predefined format checks and geoscientific sanity constraints, including bounded non-negative activity values and seasonally reasonable vegetation behavior. Valid responses are aggregated using the median. The resulting prior tables are fixed before model training and remain unchanged thereafter.

The language model is therefore used to organize process knowledge rather than to simulate carbon fluxes or provide direct \XCOtwo{} targets. Its role is not to replace observational or model-derived flux products, but to provide a scalable way to construct a globally consistent process-prior representation from dispersed process-response knowledge.

\subsection{Pixel-Wise Prior Instantiation}
The fixed prior tables are converted into spatially aligned process fields for each reconstruction date. Climate zone and vegetation type determine the baseline biospheric context of each pixel, emission level determines the anthropogenic prior, and vegetation-dependent seasonal factors modulate the biospheric fields through time. The resulting maps represent normalized relative activity associated with biospheric uptake, ecosystem respiration, and anthropogenic emissions.

These fields do not prescribe the reconstructed \XCOtwo{} value. Instead, they provide structured context that the neural network may use selectively through the knowledge adapter.

\subsection{Multi-Directional Mamba Backbone and Knowledge Adapter}
The complete 0.25$^\circ$ predictor fields are first encoded into latent spatial features. To capture long-range spatial dependencies, the encoded features are processed along four spatial scan directions, including left-to-right, right-to-left, top-to-bottom, and bottom-to-top. A shared Mamba module is reused across these directions to limit parameter redundancy and reduce sensitivity to a single scan order. The directional representations are then combined to form the backbone feature representation. (Gu \& Dao, 2023)

The structured process priors are projected into the latent feature space and spatially aligned with the backbone features. The knowledge adapter integrates the two representations through feature interaction and discrepancy modeling, and generates a process-aware feature update. A spatial gate controls the strength of prior injection at each grid cell, allowing the model to adaptively exploit process information while suppressing it where the data-driven representation provides stronger evidence. The resulting update is added to the backbone representation through a residual connection, preserving the original features while introducing structured process context.

The fused representation is subsequently decoded at two spatial scales. A coarse decoder produces the 0.25$^\circ$ reconstruction, while a fine-resolution decoder generates the final 0.05$^\circ$ \XCOtwo{} field. Both outputs are formulated as residual corrections to the CAMS background field. In addition, the fine-resolution branch predicts a spatially varying log-variance field used in the uncertainty-aware training objective described in Section 4.5.

\subsection{Training Objective}
GeoPrior-Mamba is trained using sparse OCO-2 supervision together with spatial regularization. Reconstruction losses are evaluated only at grid cells with valid OCO-2 observations on the 0.25$^\circ$ and 0.05$^\circ$ grids. The model predicts residual corrections relative to the CAMS background field. In addition to the masked reconstruction loss, we use a Gaussian negative log-likelihood term for the predicted log variance and spatial regularization terms that constrain the smoothness and local gradients of the residual field.

The overall training objective is
\begin{equation}
\mathcal{L}_{\mathrm{total}} = \mathcal{L}_{\mathrm{rec}} + \lambda_{\mathrm{nll}}\mathcal{L}_{\mathrm{nll}} + \lambda_{\mathrm{lap}}\mathcal{L}_{\mathrm{lap}} + \lambda_{\mathrm{grad}}\mathcal{L}_{\mathrm{grad}}.
\end{equation}
Here, $\mathcal{L}_{\mathrm{rec}}$ is the masked reconstruction loss, $\mathcal{L}_{\mathrm{nll}}$ denotes the uncertainty-aware Gaussian negative log-likelihood, and $\mathcal{L}_{\mathrm{lap}}$ and $\mathcal{L}_{\mathrm{grad}}$ regularize the spatial smoothness and local gradients of the predicted residual, respectively.

\begin{figure}[t]
\centering
\includegraphics[width=0.55\textwidth]{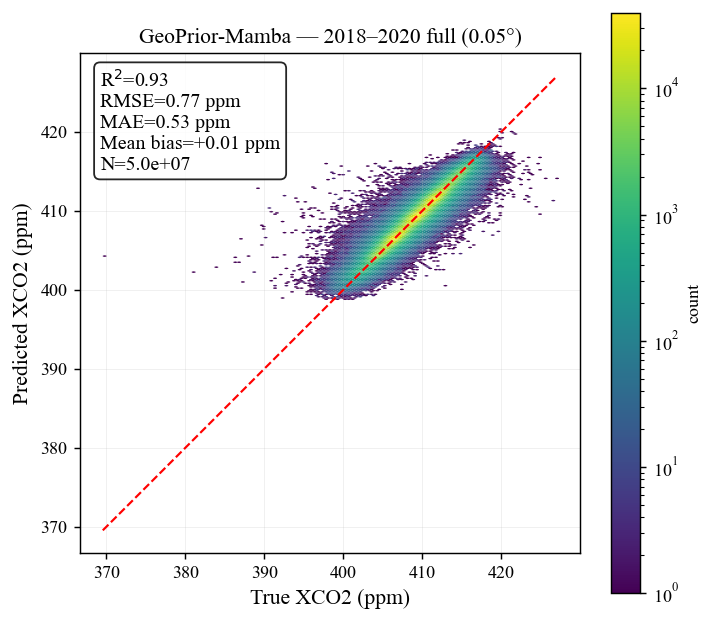}
\caption{Comparison between 0.05$^\circ$ reconstructed \XCOtwo{} and gridded OCO-2 reference \XCOtwo{} during 2018--2020.}
\label{fig:scatter}
\end{figure}

\section{Experimental Setup}
\subsection{Data Split and Evaluation}
We use OCO-2 observations from 2018--2020 and apply a date-based split so that all valid observations from held-out reconstruction dates are excluded from training. This evaluates the ability of the model to reconstruct \XCOtwo{} on unseen dates rather than merely interpolate among observations from the same day.

Performance is evaluated only at valid held-out OCO-2 grid cells using MSE, RMSE, MAE, bias, and R$^2$. Independent TCCON observations are collocated with the reconstructed fields in space and time for external validation. (Wunch et al., 2011)

\subsection{Baselines and Ablations}
We compare GeoPrior-Mamba with CAMS background interpolation and Trans-XCO2 under the same split and observation mask. We also evaluate a knowledge-free Mamba variant to isolate the contribution of the structured prior branch. Additional ablations remove the 0.25$^\circ$ branch and the NLL term. (Meng et al., 2026)

\section{Results}
\subsection{Main Reconstruction Performance}
As shown in Figure~\ref{fig:scatter}, the reconstructed \XCOtwo{} agrees well with the gridded OCO-2 reference observations over the full 2018--2020 evaluation period. GeoPrior-Mamba achieves an RMSE of 0.77 ppm, an MAE of 0.53 ppm, an MSE of 0.60 ppm$^2$, a near-zero bias, and an R$^2$ of 0.93. The close correspondence between reconstructed and observed \XCOtwo{} indicates that the model captures most of the observed variability while maintaining low systematic error.

\begin{table}[t]
\centering
\includegraphics[width=0.78\textwidth]{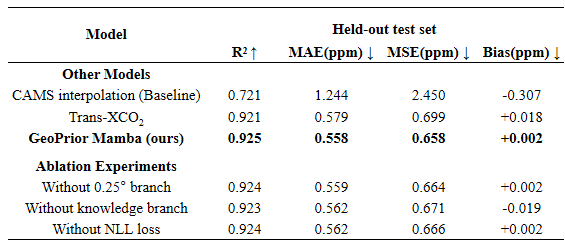}
\caption{Main reconstruction performance and ablation results on the held-out test set.}
\label{tab:heldout}
\end{table}

\begin{table}[t]
\centering
\includegraphics[width=0.80\textwidth]{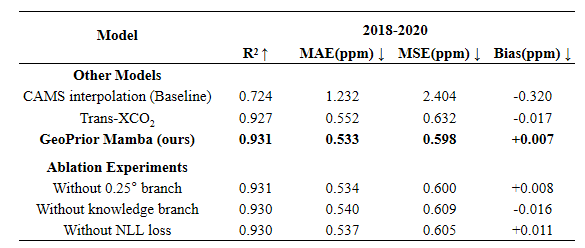}
\caption{Main reconstruction performance and ablation results during 2018--2020.}
\label{tab:fullperiod}
\end{table}

Table~\ref{tab:heldout} summarizes the main reconstruction performance on the held-out test set. GeoPrior-Mamba achieves the best overall performance among the compared methods, with an R$^2$ of 0.925, an MAE of 0.558 ppm, an MSE of 0.658 ppm$^2$, and a near-zero bias of +0.002 ppm. The corresponding RMSE is 0.811 ppm. Relative to CAMS background interpolation, GeoPrior-Mamba reduces RMSE from 1.565 to 0.811 ppm, corresponding to a reduction of approximately 48.2\%, while increasing R$^2$ from 0.721 to 0.925.

GeoPrior-Mamba also improves over Trans-XCO2 under the same evaluation protocol, reducing MSE from 0.699 to 0.658 ppm$^2$ and RMSE from 0.836 to 0.811 ppm. Although the improvement over Trans-XCO2 is smaller than that over the CAMS background, it is obtained relative to a substantially stronger learning-based baseline.

Table~\ref{tab:fullperiod} further reports the corresponding performance over the 2018--2020 period. The results show a consistent pattern, with GeoPrior-Mamba achieving an R$^2$ of 0.931 and an MSE of 0.598 ppm$^2$, compared with 0.927 and 0.632 ppm$^2$ for Trans-XCO2, respectively.

\subsection{Independent Validation with TCCON}
\begin{table}[t]
\centering
\includegraphics[width=0.96\textwidth]{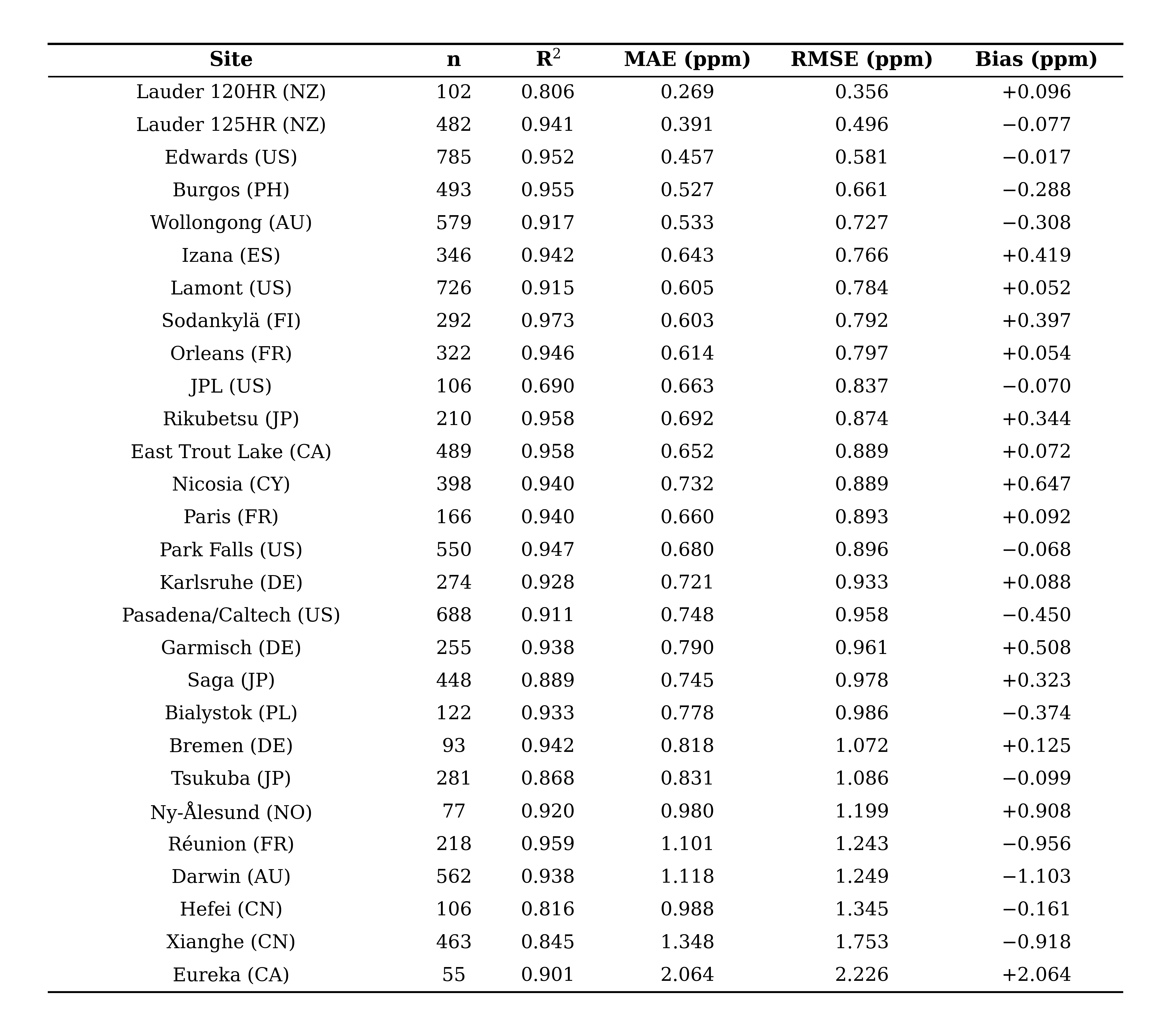}
\caption{Site-level validation against TCCON observations during 2018--2020.}
\label{tab:tccon}
\end{table}

Table~\ref{tab:tccon} summarizes the site-level validation against independent TCCON observations. The reconstructed \XCOtwo{} fields show broad consistency with ground-based column \COtwo{} measurements. R$^2$ exceeds 0.90 at most evaluated sites, and RMSE is below 1 ppm at many stations, indicating that the reconstruction captures the temporal variability of \XCOtwo{} across a wide range of locations. (Wunch et al., 2011)

Larger errors are observed at several sites, including Eureka and Xianghe. These deviations may reflect local atmospheric heterogeneity, representativeness differences between point-based TCCON measurements and 0.05$^\circ$ gridded reconstructions, and limited collocation samples at some locations. Overall, the TCCON comparison provides an independent evaluation beyond the OCO-2 observations used for model assessment.

\subsection{Contribution of the Structured Prior Module}
As shown in Tables~\ref{tab:heldout} and~\ref{tab:fullperiod}, the ablation results exhibit a consistent pattern across the held-out test set and the 2018--2020 evaluation period. Removing the knowledge-prior branch produces the largest degradation among the tested components. On the held-out test set, MSE increases from 0.658 to 0.671 and R$^2$ decreases from 0.925 to 0.923, while the bias shifts from +0.002 to -0.019 ppm. A similar pattern is observed over 2018--2020, where removing the knowledge-prior branch increases MSE from 0.598 to 0.609 and reduces R$^2$ from 0.931 to 0.930.

Removing the 0.25$^\circ$ branch results in a smaller degradation, while excluding the NLL term also slightly increases reconstruction error in both evaluation settings. These results indicate that the Mamba backbone provides a strong reconstruction foundation, while the structured prior module contributes an additional and consistent improvement in accuracy and bias correction.

\subsection{Convergence Behavior}
\begin{figure}[t]
\centering
\includegraphics[width=0.72\textwidth]{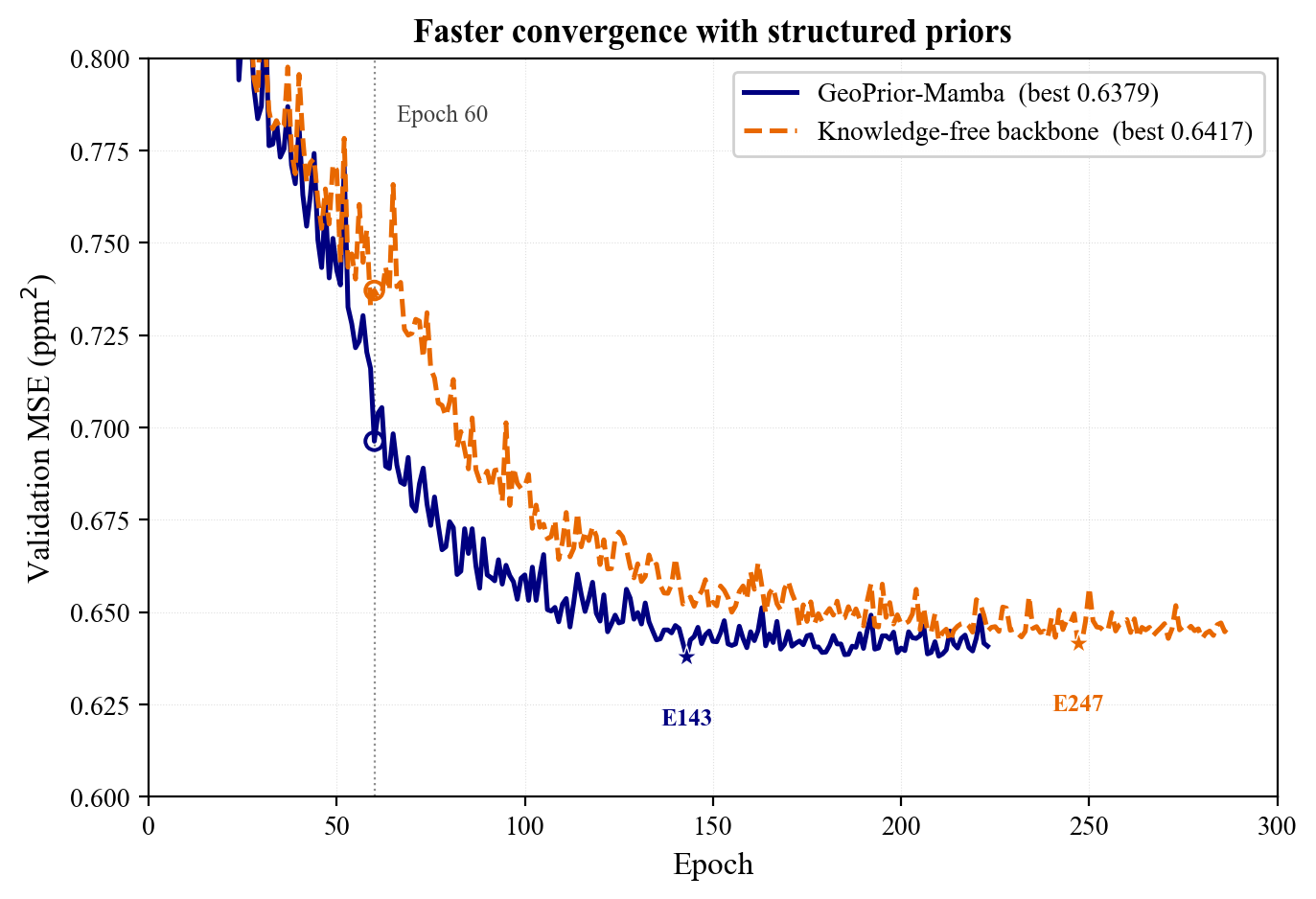}
\caption{Validation MSE curves of GeoPrior-Mamba and the knowledge-free Mamba backbone. GeoPrior-Mamba reaches the best validation level of the knowledge-free backbone substantially earlier.}
\label{fig:conv}
\end{figure}

Figure~\ref{fig:conv} compares the validation trajectories of GeoPrior-Mamba and the knowledge-free Mamba backbone. GeoPrior-Mamba reaches a validation MSE of 0.6379 at epoch 143, which is already lower than the best validation MSE achieved by the knowledge-free backbone (0.6417). The knowledge-free model requires 247 epochs to reach a comparable validation level, corresponding to 104 fewer optimization epochs, or approximately 42\% fewer epochs for GeoPrior-Mamba.

The advantage is particularly evident during the intermediate stage of training. Around epoch 60, the prior-enhanced model reaches a validation MSE of 0.6963, whereas the knowledge-free backbone remains at 0.7328. Although the gap gradually narrows as training proceeds, GeoPrior-Mamba still converges to a lower validation error. This behavior suggests that the structured priors provide useful optimization guidance under sparse supervision.

\subsection{Spatial Characteristics of the Reconstruction}
\begin{figure}[t]
\centering
\includegraphics[width=0.72\textwidth]{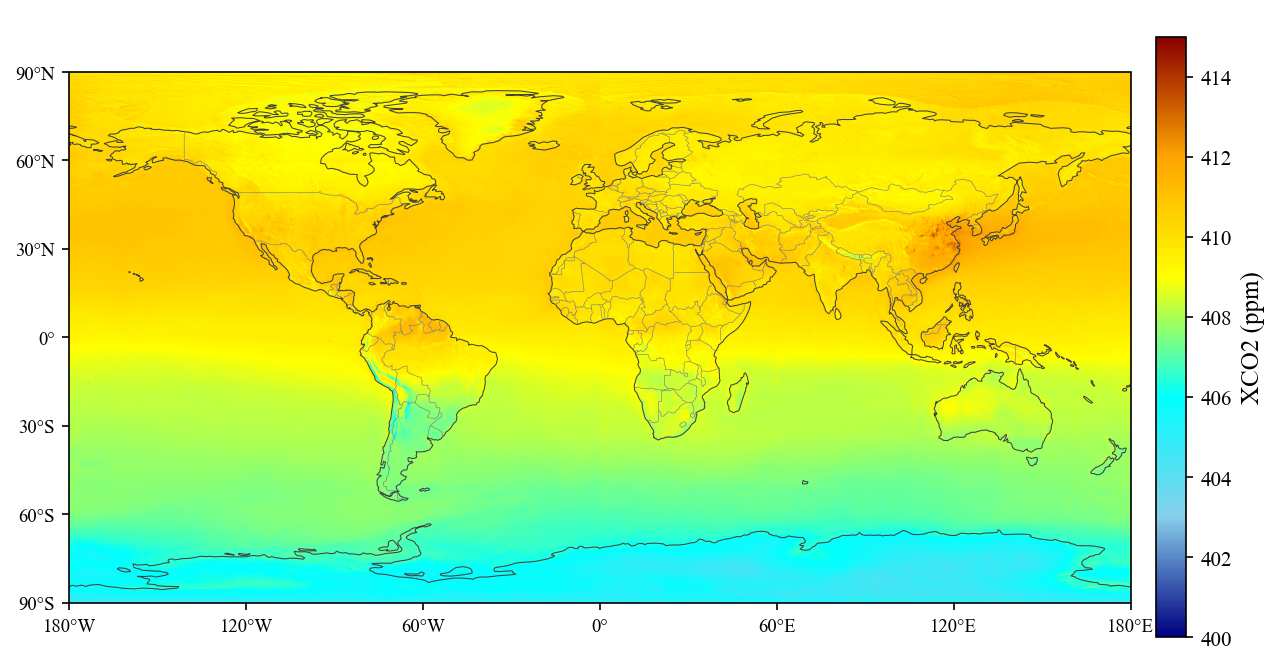}
\caption{Spatial distribution of the multi-year mean \XCOtwo{} during 2018--2020.}
\label{fig:meanmap}
\end{figure}

\begin{figure}[t]
\centering
\includegraphics[width=0.98\textwidth]{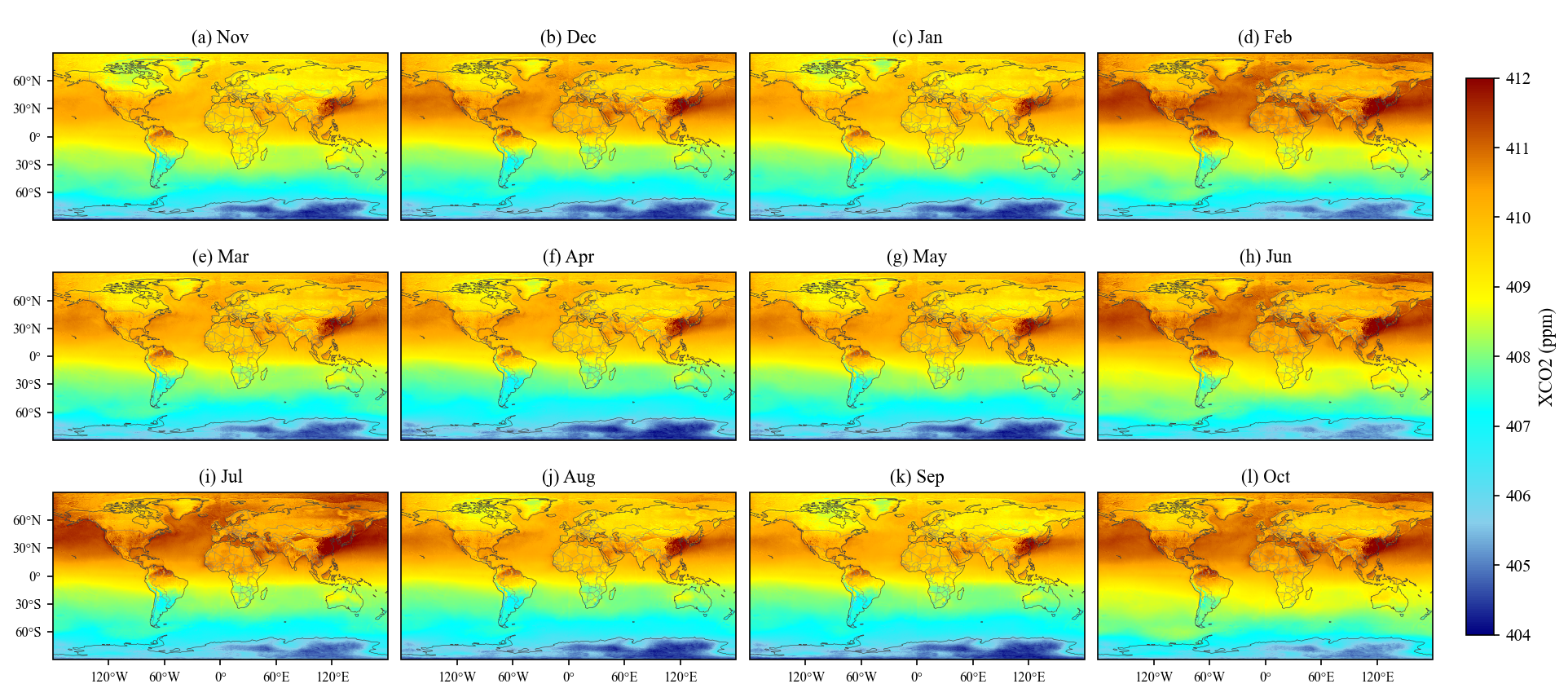}
\caption{Global distributions of monthly mean \XCOtwo{}. Panels (a--l) correspond to November through October, arranged from left to right and top to bottom.}
\label{fig:monthly}
\end{figure}

As shown in Figure~\ref{fig:meanmap}, the multi-year mean reconstruction for 2018--2020 captures the dominant large-scale structure of atmospheric \XCOtwo{}, including higher concentrations in the Northern Hemisphere and lower concentrations toward the Southern Hemisphere. Regional spatial heterogeneity is also evident over major continental regions.

Beyond the large-scale background pattern, the reconstructed fields retain finer regional contrasts associated with heterogeneous surface and atmospheric conditions. These spatial variations are particularly evident over densely populated and strongly vegetated continental regions, where anthropogenic emissions and biospheric activity contribute to more pronounced local gradients in \XCOtwo{}. (Hakkarainen et al., 2016; Lei et al., 2021)

Figure~\ref{fig:monthly} shows the multi-year monthly mean \XCOtwo{} reconstructed by GeoPrior-Mamba over 2018--2020. A clear seasonal cycle is evident, particularly over the Northern Hemisphere. \XCOtwo{} generally increases from boreal autumn into winter and remains relatively high from late winter through spring. Elevated concentrations are most apparent during approximately February--May, especially over the northern mid- and high-latitude continents. \XCOtwo{} subsequently decreases during the boreal growing season, with lower concentrations appearing over northern continental regions in July--August, before increasing again from September toward winter. Despite these seasonal variations, a persistent interhemispheric gradient is maintained throughout the year, with generally higher \XCOtwo{} in the Northern Hemisphere and lower values over the Southern Hemisphere and Southern Ocean.

The seasonal evolution is broadly consistent with the annual cycle of terrestrial carbon exchange in the Northern Hemisphere. Enhanced vegetation activity during the growing season is associated with stronger atmospheric \COtwo{} uptake and contributes to the summertime reduction in \XCOtwo{}, whereas weaker photosynthetic uptake outside the growing season is accompanied by a gradual recovery of atmospheric \COtwo{} concentrations. The seasonal contrast is particularly evident over Eurasia and North America, where the reconstructed fields show pronounced summertime drawdown relative to the preceding spring months. Regional enhancements also remain visible over densely populated continental regions, superimposed on the large-scale seasonal and interhemispheric patterns. (Graven et al., 2013; Zeng et al., 2014)

\section{Discussion and Limitations}
GeoPrior-Mamba differs from standard learning-based \XCOtwo{} reconstruction primarily in how process knowledge is represented. The language model is not used as a predictor and does not generate \XCOtwo{} fields. Instead, it is used offline to organize broad geoscientific knowledge into fixed relative-activity priors that are spatially instantiated and then interpreted adaptively by the reconstruction network. This separation preserves deterministic numerical inference while providing an explicit pathway for source--sink process knowledge to enter the model.

The current experiments indicate that the prior branch provides a measurable but moderate improvement in final reconstruction error and a clearer improvement in convergence behavior. These results support describing the priors as structured optimization guidance, rather than claiming that they dominate predictive accuracy. The stronger comparison is therefore between the knowledge-free Mamba backbone and GeoPrior-Mamba, while the comparison with Trans-XCO2 reflects the combined effect of backbone and prior design. Importantly, the present experiments evaluate the downstream utility of these LLM-elicited priors rather than establishing the empirical equivalence of individual prior values to directly measured process quantities. This suggests that language models may serve as a practical knowledge interface for constructing usable global process priors when such process-response information is difficult to assemble consistently.

Several limitations remain. First, the model is primarily data-driven and does not explicitly solve atmospheric transport equations or enforce \COtwo{} mass conservation. The reconstructed fields should therefore not be interpreted as dynamically closed simulations. Second, the language-model-induced priors are simplified representations of relative process activity rather than physical carbon fluxes. Their values depend on the prompt design, aggregation procedure, and selected process categories. Finally, evaluation over longer periods and additional observational regimes is needed to assess generalization beyond 2018--2020.

\section{Conclusion}
We introduced GeoPrior-Mamba, a framework that converts offline language-model knowledge into deterministic, spatially instantiated process priors for sparse \XCOtwo{} reconstruction. The priors encode biospheric uptake, ecosystem respiration, and anthropogenic emission context and are adaptively integrated into a multi-directional Mamba backbone. On OCO-2 reconstruction from 2018--2020, GeoPrior-Mamba improves over CAMS and Trans-XCO2 under the same evaluation protocol and provides a measurable gain over a knowledge-free Mamba backbone. The prior-enhanced model also reaches comparable validation performance in substantially fewer optimization epochs, while independent TCCON evaluation supports the broad consistency of the reconstructed fields. These results suggest that language-model knowledge can be useful for numerical geoscientific learning when it is converted into fixed, constrained process representations rather than used directly in the prediction loop. Future work should incorporate transport- or mass-conservation-related constraints and evaluate generalization across longer periods and other sparse geoscientific reconstruction tasks.

\section*{References}
\small
F. P. Antezana Lopez, G. Zhou, G. Jing, K. Zhang, L. Chen, L. Chen, and Y. Tan. Global daily column average CO2 at 0.1$^\circ$ $\times$ 0.1$^\circ$ spatial resolution integrating OCO-3, GOSAT, CAMS with EOF and deep learning, 2025. Sci. Data, vol. 12, Art. no. 268, doi: 10.1038/s41597-024-04135-w.\par
S. Bhattacharjee and J. Chen. Prediction of satellite-based column CO2 concentration by combining emission inventory and LULC information, 2020. IEEE Trans. Geosci. Remote Sens., vol. 58, no. 12, pp. 8285--8300, doi: 10.1109/TGRS.2020.2985047.\par
B. Byrne, D. F. Baker, S. Basu, M. Bertolacci, K. W. Bowman, D. Carroll, et al. National CO2 budgets (2015--2020) inferred from atmospheric CO2 observations in support of the global stocktake, 2023. Earth Syst. Sci. Data, vol. 15, no. 2, pp. 963--1004, doi: 10.5194/essd-15-963-2023.\par
J. Chen, R. Hu, L. Chen, Z. Liao, L. Che, and T. Li. Multi-sensor integrated mapping of global XCO2 from 2015 to 2021 with a local random forest model, 2024. ISPRS J. Photogramm. Remote Sens., vol. 208, pp. 107--120, doi: 10.1016/j.isprsjprs.2024.01.009.\par
W. Cheng, L. Dan, X. Deng, X. Huang, et al. Global monthly gridded atmospheric carbon dioxide concentrations under the historical and future scenarios, 2022. Sci. Data, vol. 9, no. 1, Art. no. 83, doi: 10.1038/s41597-022-01196-7.\par
A. Eldering, P. O. Wennberg, D. Crisp, et al. The Orbiting Carbon Observatory-2 early science investigations of regional carbon dioxide fluxes, 2017. Science, vol. 358, no. 6360, Art. no. eaam5745, doi: 10.1126/science.aam5745.\par
P. Friedlingstein, M. O'Sullivan, M. W. Jones, et al. Global carbon budget 2022, 2022. Earth Syst. Sci. Data, vol. 14, no. 11, pp. 4811--4900, doi: 10.5194/essd-14-4811-2022.\par
Z. Gao, Y. Jiang, J. He, et al. Spatiotemporal variation analysis of global XCO2 concentration during 2010--2020 based on DINEOF-BME framework and wavelet function, 2023. Sci. Total Environ., vol. 892, Art. no. 164750, doi: 10.1016/j.scitotenv.2023.164750.\par
H. D. Graven, R. F. Keeling, S. C. Piper, et al. Enhanced seasonal exchange of CO2 by northern ecosystems since 1960, 2013. Science, vol. 341, no. 6150, pp. 1085--1089, doi: 10.1126/science.1239207.\par
A. Gu and T. Dao. Mamba: Linear-time sequence modeling with selective state spaces, 2023. arXiv:2312.00752.\par
X. Guan, Z. Sun, D. Chu, et al. Long-term (2000--2020) global 0.05$^\circ$ continuous atmospheric carbon dioxide mapping combining OCO-2 observations and model simulations, 2024. Sci. Total Environ., vol. 957, Art. no. 177051, doi: 10.1016/j.scitotenv.2024.177051.\par
J. Hakkarainen, I. Ialongo, and J. Tamminen. Direct space-based observations of anthropogenic CO2 emission areas from OCO-2, 2016. Geophys. Res. Lett., vol. 43, no. 21, pp. 11400--11406, doi: 10.1002/2016GL070885.\par
D. M. Hammerling, A. M. Michalak, and S. R. Kawa. Mapping of CO2 at high spatiotemporal resolution using satellite observations: Global distributions from OCO-2, 2012. J. Geophys. Res. Atmos., vol. 117, no. D6, doi: 10.1029/2011JD017015.\par
C. He, M. Ji, T. Li, et al. Deriving full-coverage and fine-scale XCO2 across China based on OCO-2 satellite retrievals and CarbonTracker output, 2022. Geophys. Res. Lett., vol. 49, no. 12, Art. no. e2022GL098435, doi: 10.1029/2022GL098435.\par
Q. He, Y. Liu, Z. Li, et al. Spatiotemporal dynamics of atmospheric CO2 across China revealed by long-term, high-resolution satellite-derived data, 2026. EGUsphere, pp. 1--23, doi: 10.5194/egusphere-2025-5647.\par
H. Hersbach, B. Bell, P. Berrisford, et al. The ERA5 global reanalysis, 2020. Q. J. R. Meteorol. Soc., vol. 146, no. 730, pp. 1999--2049, doi: 10.1002/qj.3803.\par
Y. Hua, X. Zhao, W. Sun, et al. Satellite-based reconstruction of atmospheric CO2 concentration over China using a hybrid CNN and spatiotemporal kriging model, 2024. Remote Sens., vol. 16, no. 13, Art. no. 2433, doi: 10.3390/rs16132433.\par
X. Huang, H. Yang, Q. Lv, et al. A full-coverage daily XCO2 dataset in China from 2015 to 2020 based on DSC-DF-LGB, 2024. Earth Syst. Sci. Data Discuss., pp. 1--22, doi: 10.5194/essd-2024-371.\par
A. Huete, K. Didan, T. Miura, E. P. Rodriguez, X. Gao, and L. G. Ferreira. Overview of the radiometric and biophysical performance of the MODIS vegetation indices, 2002. Remote Sens. Environ., vol. 83, no. 1--2, pp. 195--213, doi: 10.1016/S0034-4257(02)00096-2.\par
S. Hwang, H. Choi, Y. Kang, et al. Reconstructing long-term (2003--2019) global high-resolution XCO2: Bridging observational gaps with machine learning, 2026. GISci. Remote Sens., vol. 63, no. 1, Art. no. 2627042, doi: 10.1080/15481603.2026.2627042.\par
A. Inness, M. Ades, A. Agust\'i-Panareda, et al. The CAMS reanalysis of atmospheric composition, 2019. Atmos. Chem. Phys., vol. 19, pp. 3515--3556, doi: 10.5194/acp-19-3515-2019.\par
C. Jin, Y. Xue, X. Jiang, et al. A long-term global XCO2 dataset: Ensemble of satellite products, 2022. Atmos. Res., vol. 279, Art. no. 106385, doi: 10.1016/j.atmosres.2022.106385.\par
R. Lei, S. Feng, A. Danjou, et al. Fossil fuel CO2 emissions over metropolitan areas from space: A multi-model analysis of OCO-2 data over Lahore, Pakistan, 2021. Remote Sens. Environ., vol. 264, Art. no. 112625, doi: 10.1016/j.rse.2021.112625.\par
J. Li, K. Jia, X. Wei, et al. High-spatiotemporal resolution mapping of spatiotemporally continuous atmospheric CO2 concentrations over the global continent, 2022. Int. J. Appl. Earth Obs. Geoinf., vol. 108, Art. no. 102743, doi: 10.1016/j.jag.2022.102743.\par
J. Li, Z. Zhang, T. Li, et al. Global daily seamless XCO2 mapping (2016--2020): Spatio-temporal trends and variations during wildfire events, 2026. Int. J. Appl. Earth Obs. Geoinf., vol. 146, Art. no. 105092, doi: 10.1016/j.jag.2026.105092.\par
T. Li, J. Wu, and T. Wang. Generating daily high-resolution and full-coverage XCO2 across China from 2015 to 2020 based on OCO-2 and CAMS data, 2023. Sci. Total Environ., vol. 893, Art. no. 164921, doi: 10.1016/j.scitotenv.2023.164921.\par
J. Liang, J. Cao, G. Sun, et al. SwinIR: Image restoration using swin transformer, 2021. in Proc. IEEE/CVF Int. Conf. Comput. Vis. (ICCV), pp. 1833--1844.\par
W. Liu, R. Li, J. Cao, et al. Mapping high-resolution XCO2 concentrations in China from 2015 to 2020 based on spatiotemporal ensemble learning model, 2024. Ecol. Inform., vol. 83, Art. no. 102806, doi: 10.1016/j.ecoinf.2024.102806.\par
Z. Meng, J. Zheng, Y. Zhu, and H. Fu. Global daily XCO2 reconstruction at 0.05$^\circ$ resolution for 2021--2024 using satellite and reanalysis data, 2026. IEEE Trans. Geosci. Remote Sens., doi: 10.1109/TGRS.2026.3716990.\par
T. Oda and S. Maksyutov. A very high-resolution (1 km $\times$ 1 km) global fossil fuel CO2 emission inventory derived using a point source database and satellite observations of nighttime lights, 2011. Atmos. Chem. Phys., vol. 11, pp. 543--556, doi: 10.5194/acp-11-543-2011.\par
G. Pan, Y. Xu, and J. Ma. The potential of CO2 satellite monitoring for climate governance: A review, 2021. J. Environ. Manage., vol. 277, Art. no. 111423, doi: 10.1016/j.jenvman.2020.111423.\par
S. Pandey, J. B. Miller, S. Basu, et al. Toward low-latency estimation of atmospheric CO2 growth rates using satellite observations: Evaluating sampling errors of satellite and in situ observing approaches, 2024. AGU Adv., vol. 5, no. 4, Art. no. e2023AV001145, doi: 10.1029/2023AV001145.\par
Y. Qu, X. Shi, Y. Fan, et al. Reconstructing two-decadal global daily high-resolution XCO2 records using a hybrid Transformer--BiLSTM model, 2026. Earth Syst. Sci. Data Discuss., pp. 1--35, doi: 10.5194/essd-2026-50.\par
M. Raissi, P. Perdikaris, and G. E. Karniadakis. Physics-informed neural networks: A deep learning framework for solving forward and inverse problems involving nonlinear partial differential equations, 2019. J. Comput. Phys., vol. 378, pp. 686--707, doi: 10.1016/j.jcp.2018.10.045.\par
M. Reuter, M. Buchwitz, O. Schneising, et al. Ensemble-based satellite-derived carbon dioxide and methane column-averaged dry-air mole fraction data sets (2003--2018) for carbon and climate applications, 2020. Atmos. Meas. Tech., vol. 13, no. 2, pp. 789--819, doi: 10.5194/amt-13-789-2020.\par
M. Sheng, L. Lei, Z. C. Zeng, et al. Global land 1$^\circ$ mapping dataset of XCO2 from satellite observations of GOSAT and OCO-2 from 2009 to 2020, 2023. Big Earth Data, vol. 7, no. 1, pp. 170--190, doi: 10.1080/20964471.2022.2033149.\par
Y. Tang, J. Hu, D. Tang, et al. Spatio-temporal distribution of XCO2 concentrations in northeast China based on Downscaling-XGBoost model, 2024. in Proc. Int. Conf. Remote Sens., Mapping, Geograph. Inf. Syst. (RSMG), vol. 13402. Bellingham, WA, USA: SPIE, pp. 367--372, doi: 10.1117/12.3048860.\par
W. Tian, L. Zhang, Y. Yu, et al. Using multi-source data and time series features to construct a global terrestrial CO2 coverage by deep learning, 2024. IEEE Trans. Geosci. Remote Sens., doi: 10.1109/TGRS.2024.3462589.\par
J. Wang. Global daily 1 km gapless XCO2 (2003--2023) derived from multi-satellite observations and a spatiotemporal deep learning framework, 2026. Environ. Impact Assess. Rev., vol. 117, Art. no. 108146, doi: 10.1016/j.eiar.2025.108146.\par
Y. Wang, Q. Yuan, T. Li, et al. Seamless mapping of long-term (2010--2020) daily global XCO2 and XCH4 from the Greenhouse Gases Observing Satellite (GOSAT), Orbiting Carbon Observatory 2 (OCO-2), and CAMS global greenhouse gas reanalysis (CAMS-EGG4) with a spatiotemporally self-supervised fusion method, 2023. Earth Syst. Sci. Data, vol. 15, no. 8, pp. 3597--3622, doi: 10.5194/essd-15-3597-2023.\par
Z. Wang, C. Zhang, K. Shi, Y. Shangguan, B. Hu, X. Chen, D. Wei, S. Chen, P. M. Atkinson, and Q. Zhang. A full-coverage satellite-based global atmospheric CO2 dataset at 0.05$^\circ$ resolution from 2015 to 2021 for exploring global carbon dynamics, 2025. Earth Syst. Sci. Data, vol. 17, pp. 5355--5375, doi: 10.5194/essd-17-5355-2025.\par
C. Wu, Y. Ju, S. Yang, et al. Reconstructing annual XCO2 at a 1 km $\times$ 1 km spatial resolution across China from 2012 to 2019 based on a spatial CatBoost method, 2023. Environ. Res., vol. 236, Art. no. 116866, doi: 10.1016/j.envres.2023.116866.\par
D. Wunch, G. C. Toon, J.-F. L. Blavier, et al. The Total Carbon Column Observing Network, 2011. Philos. Trans. R. Soc. A, vol. 369, no. 1943, pp. 2087--2112, doi: 10.1098/rsta.2010.0240.\par
H. Yang, T. Li, J. Wu, et al. Inter-comparison and evaluation of global satellite XCO2 products, 2025. Geo-Spat. Inf. Sci., vol. 28, no. 1, pp. 131--144, doi: 10.1080/10095020.2023.2252017.\par
N. Zeng, F. Zhao, G. J. Collatz, et al. Agricultural Green Revolution as a driver of increasing atmospheric CO2 seasonal amplitude, 2014. Nature, vol. 515, no. 7527, pp. 394--397, doi: 10.1038/nature13893.\par
L. Zhang, F. Jiang, W. He, et al. A robust estimate of continental-scale terrestrial carbon sinks using GOSAT XCO2 retrievals, 2023a. Geophys. Res. Lett., vol. 50, no. 6, Art. no. e2023GL102815, doi: 10.1029/2023GL102815.\par
L. Zhang, T. Li, J. Wu, et al. Global estimates of gap-free and fine-scale CO2 concentrations during 2014--2020 from satellite and reanalysis data, 2023b. Environ. Int., vol. 178, Art. no. 108057, doi: 10.1016/j.envint.2023.108057.\par
M. Zhang and G. Liu. Mapping contiguous XCO2 by machine learning and analyzing the spatio-temporal variation in China from 2003 to 2019, 2023. Sci. Total Environ., vol. 858, Art. no. 159588, doi: 10.1016/j.scitotenv.2022.159588.\par
Y. Zhang, X. Chen, B. Jin, S. Wang, S. Ji, W. Wang, and J. Han. A comprehensive survey of scientific large language models and their applications in scientific discovery, 2024. in Proc. 2024 Conf. Empirical Methods in Natural Language Processing (EMNLP), pp. 8783--8817, doi: 10.18653/v1/2024.emnlp-main.498.\par
J. Zscheischler, M. D. Mahecha, V. Avitabile, et al. Reviews and syntheses: An empirical spatiotemporal description of the global surface--atmosphere carbon fluxes: Opportunities and data limitations, 2017. Biogeosciences, vol. 14, no. 15, pp. 3685--3703, doi: 10.5194/bg-14-3685-2017.\par
\end{document}